\documentclass[10pt,twocolumn,letterpaper]{article}

\usepackage{cvpr}              

\definecolor{cvprblue}{rgb}{0.21,0.49,0.74}
\usepackage[pagebackref,breaklinks,colorlinks,allcolors=cvprblue]{hyperref}
\usepackage{multirow}
\usepackage{gensymb}
\usepackage{comment}
\usepackage{graphicx}
\usepackage{tablefootnote}

\newcommand{\R}[0]{\ensuremath{\mathcal{R}}}
\newcommand{\Q}[0]{\ensuremath{\mathcal{Q}}}
\newcommand{\Mat}{\boldsymbol}

\def\paperID{3915}
\def\confName{CVPR}
\def\confYear{2025}

\title{Dense-SfM: Structure from Motion with Dense Consistent Matching}

\author{JongMin Lee\\
Seoul National University\\
{\tt\small jongmin.cv@gmail.com}
\and
Sungjoo Yoo\\
Seoul National University\\
{\tt\small sungjoo.yoo@gmail.com}
}

\begin{document}

\maketitle    
\begin{abstract}

We present Dense-SfM, a novel Structure from Motion (SfM) framework designed for dense and accurate 3D reconstruction from multi-view images. Sparse keypoint matching, which traditional SfM methods often rely on, limits both accuracy and point density, especially in texture-less areas. Dense-SfM addresses this limitation by integrating dense matching with a Gaussian Splatting (GS) based track extension which gives more consistent, longer feature tracks. To further improve reconstruction accuracy, Dense-SfM is equipped with a multi-view kernelized matching module leveraging transformer and Gaussian Process architectures, for robust track refinement across multi-views. Evaluations on the ETH3D and Texture-Poor SfM datasets show that Dense-SfM offers significant improvements in accuracy and density over state-of-the-art methods. Project page: \url{https://icetea-cv.github.io/densesfm/}.

\end{abstract}

\section{Introduction}


\begin{figure}
    \centering
    \resizebox{\linewidth}{!}{
    
    \raisebox{-0.5\height}
    {\includegraphics{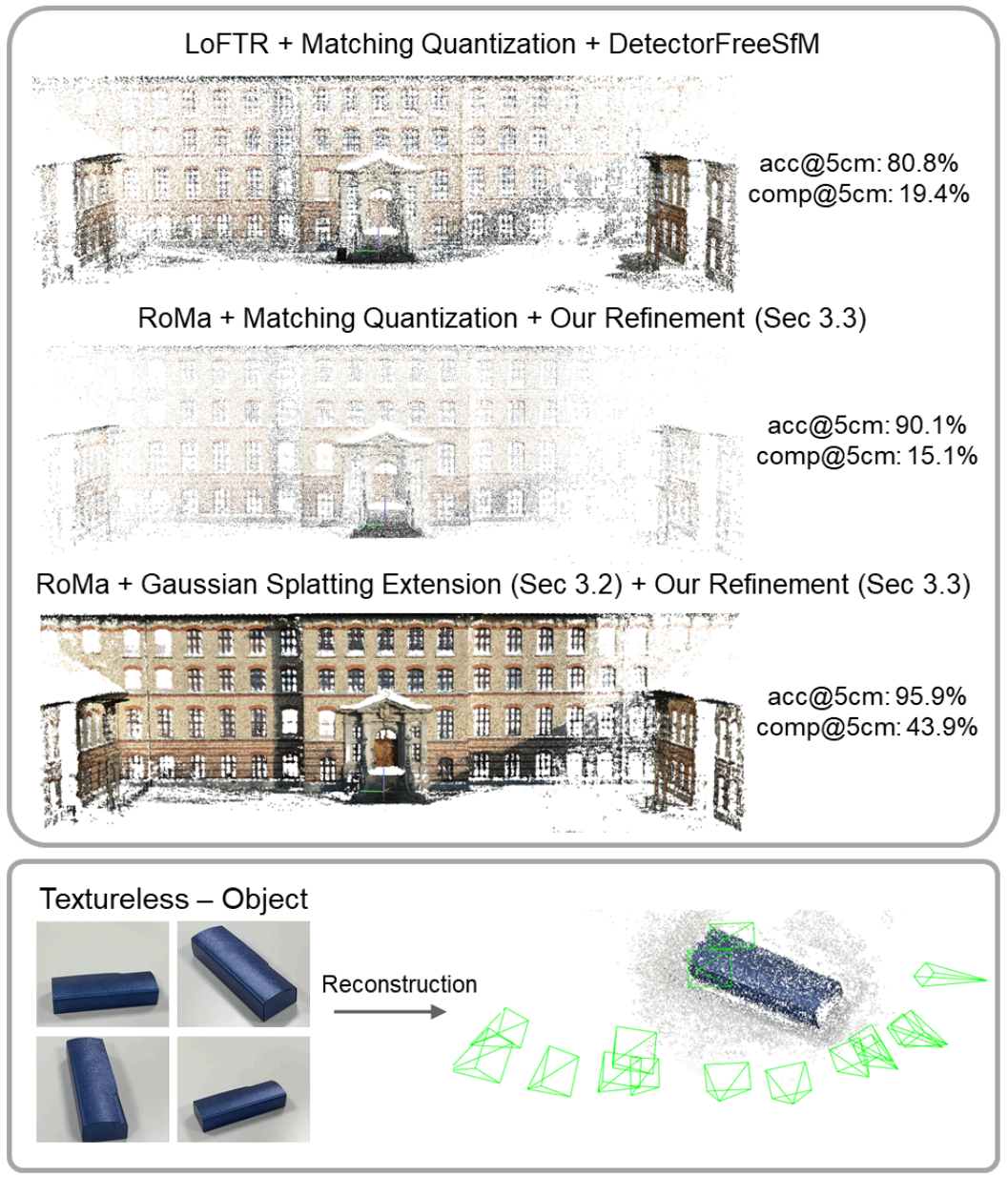}}}
    \vspace{-2mm}
    \caption{\textbf{Comparison of SfM results between quantized-matching and our matching.} 
    To solve the fragmentary track problem, DFSfM~\cite{detectorfreesfm} uses quantized matching, which can compromise the accuracy and completeness of 3D points due to quantization. In contrast, our Dense-SfM pipeline leverages Gaussian Splatting to solve the problem, offering more accurate and dense 3D points. 
    Thanks to dense matching, our pipeline can also be applied to texture-less objects, further improving the quality of 3D points and pose estimation.}
    \vspace{-5mm}
    \label{fig:intro}
\end{figure}

Structure from Motion (SfM) is a fundamental task in computer vision that enables the reconstruction of 3D structures from multi-view images of a scene. It has been a widely researched problem, focusing on reconstructing accurate camera poses and 3D point cloud of a scene, benefiting a variety of downstream tasks such as visual localization~\cite{hloc, sacreg, visual_loc_1, visual_loc_2, neural_reprojection_error}, AR/VR~\cite{robot_vision_AR, robot_vision2}, multi-view stereo~\cite{mvs1, mvs2, mvs3} and neural rendering~\cite{nerf, naive_gs, fsgs}. Over the years, several well-established methods have been developed to address the SfM pipeline~\cite{incrementalsfm, colmap, photo_tourism, openmvg, glomap}.

Recently, there has been an increasing demand for dense and accurate 3D reconstructions in various 3D vision tasks, including dense reconstruction~\cite{neus, geoneus}, object pose estimation~\cite{onepose, oneposepp, posematcher} and neural rendering~\cite{naive_gs, fsgs}. However, existing detector-based methods have limitations in achieving sufficient accuracy and density of 3D points, especially in texture-less regions where obtaining matchable keypoints is challenging, limiting the performance of downstream tasks.
With the advent of deep learning and attention mechanism~\cite{transformer}, semi-dense and dense matching methods~\cite{loftr, wang2022matchformer, Chen2022ASpanFormerDI, dkm, roma} have been introduced, enabling the reliable matches even in low-texture areas and producing denser match results than classical detector-based methods. This advancement has led to improved performance across many tasks, including homography and relative pose estimation~\cite{hpatches, scannet, megadepth}.  
However, due to their pairwise matching approach, the matching methods usually yield fragmented tracks\footnote{A keypoint is assigned a set of 2D coordinates, called feature track or track in short, on neighbor images where the keypoint is matched.} as mentioned in Detector-Free SfM (DFSfM)~\cite{detectorfreesfm}, making direct integration with existing SfM pipelines challenging. To solve this inconsistency issue, DFSfM applied quantization to the matched keypoints, and then refined the multi-view tracks to obtain accurate camera pose and point cloud. While the quantization strategy can increase track length and consistency, enhancing the multi-view refinement results, it can significantly limit the number of matches and accuracy, reducing the quality of the resulting SfM model, as exemplified in Fig.~\ref{fig:intro}.

In this paper, we propose a novel SfM framework that leverages dense matching while addressing the limitations of pairwise matching methods. As shown in Fig.~\ref{fig:intro}, our approach reconstructs a dense and accurate 3D point cloud, avoiding the inherent limitations of quantized matching. 
Specifically, our framework first performs dense matching using a dense matcher, \eg, DKM~\cite{dkm} or RoMa~\cite{roma}, to construct an initial SfM model, where most tracks are initially fragmentary, \ie, of short (mostly a pair of images) lengths.
Next, we initialize small 3D Gaussians based on the initial SfM's point cloud, integrating attributes like color and position. These Gaussians are then optimized and densified through Gaussian Splatting (GS)~\cite{naive_gs, fsgs}, filling in the scene while keeping the initial Gaussians' (SfM points') positions fixed. We then infer the visibility of each initial SfM point per image by rendering through the ray from camera poses. In case that the 3D point is visible on the image, we include the image into the track of the 3D point thereby increasing the track length—a process we refer to as \textit{Track Extension}. Note that \textit{Track Extension} enables us to resolve the problem of short tracks in pairwise matching. Lastly, we perform our proposed SfM refinement pipeline, which includes a multi-view kernelized matching module and geometric bundle adjustment, to further optimize the camera poses and 3D points. 
Our track refinement module is mainly built on two architectures, transformer and Gaussian Process~\cite{gaussian_process}, which incorporate both feature-based and positional information, resulting in notable performance gains over existing refinement modules as demonstrated in our ablation study (Tab.~\ref{tab:ablation}).

We evaluate our Dense-SfM framework on the ETH3D, Texture-Poor SfM and Image Matching Challenge (IMC) 2021 datasets, where it outperforms state-of-the-art SfM systems across multiple metrics. Thanks to the dense matching coupled with track extension via Gaussian Splatting, as well as our novel track refinement module, our framework achieves accurate camera pose estimations and dense 3D reconstructions.

\vspace{-2mm}

\paragraph{Contributions:}
\begin{itemize}
[leftmargin=*]

\item A novel SfM framework optimized for dense matchers, capable of reconstructing highly accurate and dense 3D models, even in texture-less regions.
\item A track extension method that, based on Gaussian Splatting, helps find new images on which 3D point is visible, thereby increasing track length of the 3D point.
\item A new track refinement pipeline that leverages transformer and Gaussian Process architectures to enhance the consistency of feature tracks and refine the overall reconstruction quality.
\item Extensive evaluations across diverse metrics demonstrate that our framework achieves state-of-the-art performance.

\end{itemize}

\section{Related Work}
\paragraph{Image Matching.}
Establishing 2D correspondences through feature matching is a fundamental component for SfM and visual SLAM. Traditional feature matching typically involves i) detecting keypoints~\cite{orb, sift, fast, superpoint, r2d2, d2net}, ii) computing descriptors~\cite{geodesc, disk, sosnet, lfnet, logpolar_desc, aslfeat} and iii) performing descriptor matching~\cite{superglue, oanet, s2dnet}. Despite its efficiency, it struggles in challenging scenarios such as low-textured regions, where keypoint detection is unreliable, and repetitive patterns, where matching often fails—leading to suboptimal SfM results.

To address these limitations, more recent methods bypass the need for keypoint detection and instead directly perform semi-dense~\cite{loftr, Chen2022ASpanFormerDI, wang2022matchformer} or dense~\cite{roma, dkm} matching between image pairs. These approaches benefit from global receptive fields provided by transformers~\cite{transformer} or correlations computed from CNN or ViT features~\cite{vit, oquab2023dinov2}, which allows them to achieve better performance in texture-poor areas where traditional methods often fail.

However, semi-dense and dense matching often results in fragmentary feature tracks, making it difficult to apply on SfM pipeline~\cite{Shen2022SemiDenseFM}. To solve this, DFSfM~\cite{detectorfreesfm} introduced quantization matching that merges nearby subpixel matches into grid nodes, improving track consistency. Yet, this approach can compromise the accuracy of the initial camera poses and point clouds, heavily relying on refinement modules and reducing point cloud density.

In contrast, our framework leverages Gaussian Splatting to assess visibility of 3D points. By projecting 3D points onto new images on which the points are visible, we obtain longer and consistent tracks without sacrificing the initial matching accuracy, which leads to improved refinement performance and, finally, better SfM results.

\begin{figure*}[ht!]
    \centering
    \resizebox{\linewidth}{!}{
    
    \raisebox{-0.5\height}
    {\includegraphics{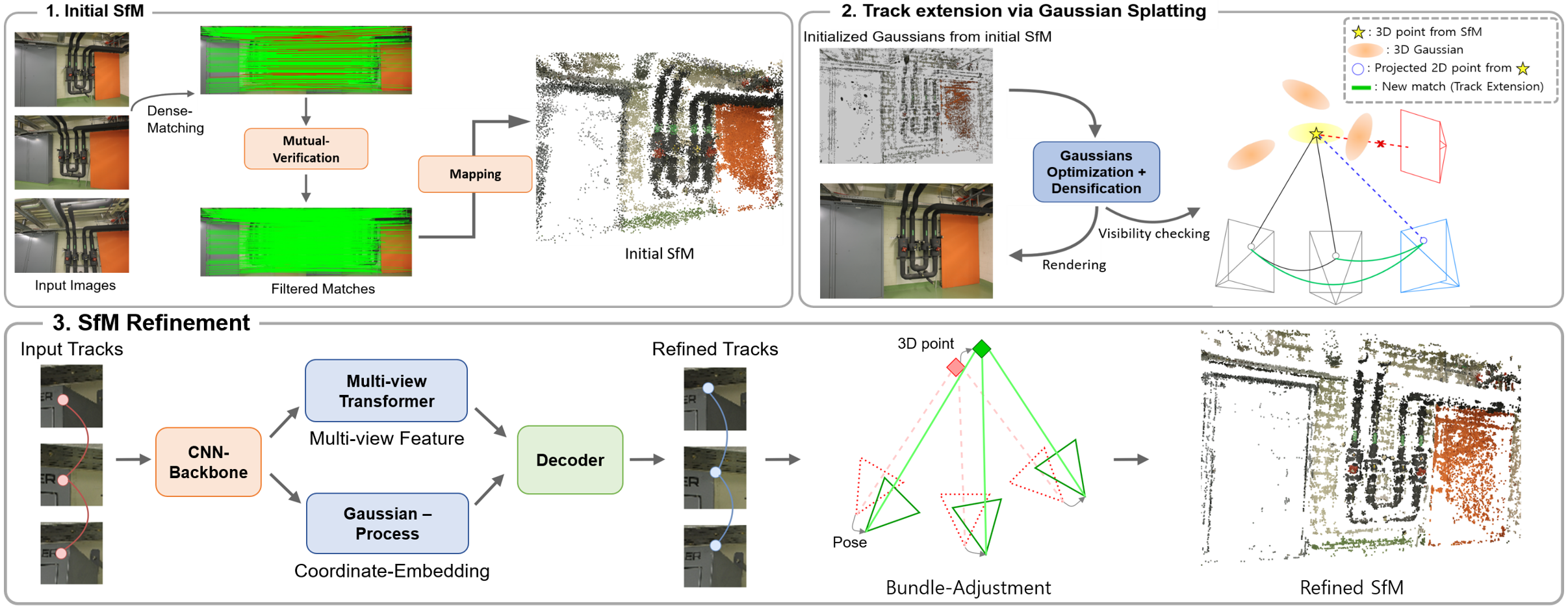}}}
    \vspace{-2mm}
    \caption{\textbf{Pipeline Overview.} From a set of images, we construct an initial SfM model using dense two-view matching, filtering unreliable matches through mutual verification. To extend track length, we project 3D points onto additional images, using a visibility filter based on Gaussian Splatting. We then refine these extended tracks with our track refinement module and perform geometric bundle adjustment to improve the accuracy of SfM model.}
    \vspace{-3mm}
    \label{fig:overview}
\end{figure*}

\paragraph{Structure From Motion.}
Feature correspondence-based SfM methods have been extensively studied~\cite{sfm1, sfm2, sfm3}. Incremental SfM~\cite{build_rome_in_a_day, photo_tourism, incrementalsfm} has become a widely adopted approach, with COLMAP~\cite{colmap} recognized as one of the most popular frameworks.

Recently, differentiable SfM methods~\cite{spann3r, vggsfm, acezero, flowmap, dust3r, mast3r-sfm} have aimed to simplify or entirely replace the traditional SfM pipeline. VGGSfM~\cite{vggsfm} manages all SfM components as fully differentiable, enabling end-to-end training. 
DUSt3R~\cite{dust3r} predicts pairwise pointmaps and refines them through optimization-based global alignment. Its successors, MASt3R~\cite{mast3r} and MASt3R-SfM~\cite{mast3r-sfm}, improve pixel matching for localization by predicting additional per-pixel features and leveraging its encoder for fast image retrieval, leading to a newly proposed SfM pipeline.



\paragraph{Multi-view Refinement.}
Inconsistent feature points across multiple views can impact the accuracy of reconstructed point cloud and camera poses in SfM. Hence, precise multi-view correspondences are essential for accurately recovering 3D structures and camera parameters in SfM pipelines. 

To address this, several previous methods have proposed multi-view keypoint refinement models based on flow~\cite{multiview_optimization_of_local_feature_geometry} or dense features~\cite{pixsfm}, which lead to significant improvements in SfM accuracy. PatchFlow~\cite{multiview_optimization_of_local_feature_geometry} refines keypoint locations using a geometric cost constrained by local flow between two views, followed by global optimization using a multi-view graph of relative feature geometry constraints. 
Similarly, PixSfM~\cite{pixsfm} refines keypoint locations through feature-metric keypoint adjustment by optimizing a direct cost over discriminative dense feature maps along the tracks.
DFSfM~\cite{detectorfreesfm}, introduces a multi-view matching module that iteratively refines the tracks obtained from the COLMAP framework. The module extracts local patches from each image and feeds them into a transformer-based network to refine the keypoint's location by enforcing multi-view consistency. 

Building on these recent advancements, we propose a novel refinement module that leverages both feature and positional information to refine the keypoint coordinates within the tracks. Instead of relying on statistical methods to select the refined tracks, our module fully regresses and learns the confidence (certainty) per track, resulting in more accurate refinement. As demonstrated in our ablation study (Tab.~\ref{tab:ablation}), our approach outperforms the existing refinement methods.

\section{Method}
\label{sec:intro}

An overview of our Dense-SfM framework is
shown in Fig.~\ref{fig:overview}. Given $N$ images \{${I_i}$\} of a scene, 
our goal is to estimate the camera poses and reconstruct the 3D point cloud representing the scene. 

We design our framework as a three-stage pipeline. First, we construct an initial SfM model using correspondences between image pairs from a two-view dense matcher (\eg, DKM~\cite{dkm} and RoMa~\cite{roma}). Next, we extend the track length of each 3D point by projecting it onto new images where it is visible, verifying visibility through Gaussian Splatting~\cite{naive_gs}. Finally, we perform iterative SfM refinement to enhance the accuracy of poses and point cloud.


\begin{figure}
    \centering
    \resizebox{0.7\linewidth}{!}{
    
    \raisebox{-0.5\height}
    {\includegraphics{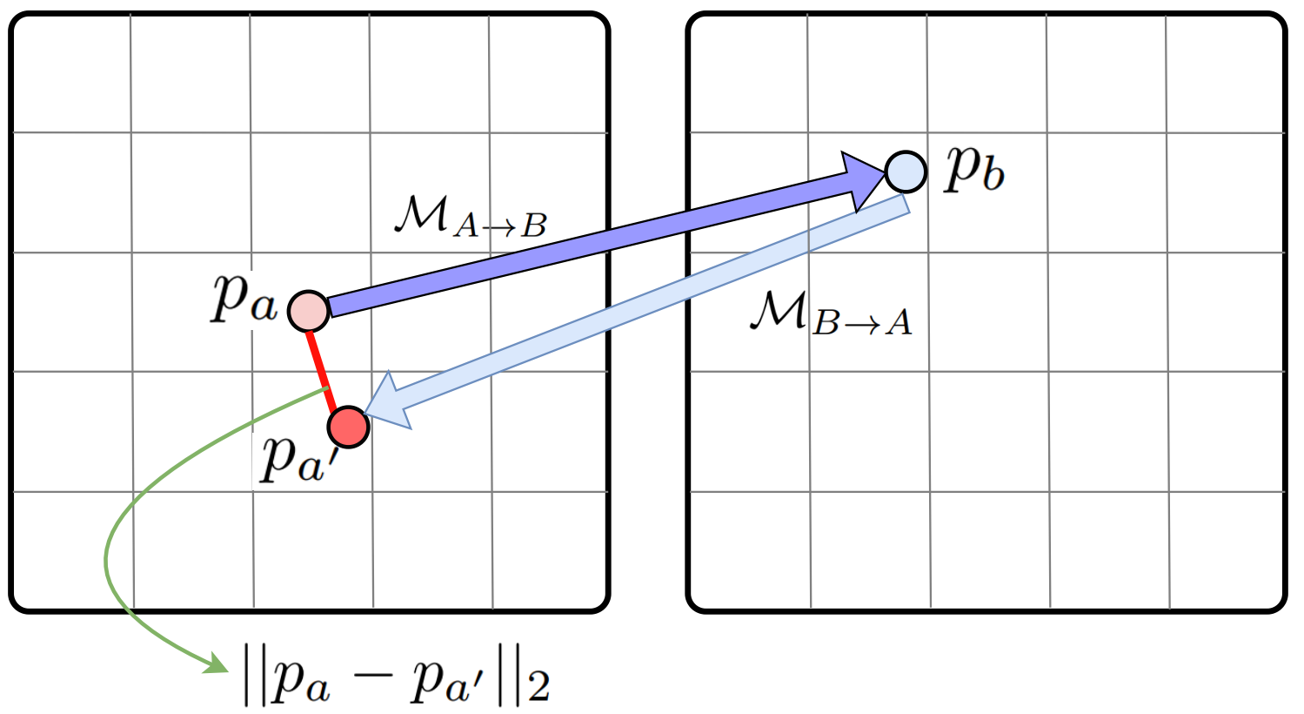}}}
    \vspace{-2mm}
    \caption{\textbf{Bidirectional verification on two-view dense matching.} The match result $p_b$ from $\mathbf{\mathcal{M}}_{A\rightarrow{} B}$ is re-used as input to $\mathbf{\mathcal{M}}_{B\rightarrow{} A}$ to estimate $p_{a'}$. We then compute the distance between $p_a$ and $p_{a'}$, where a smaller distance indicates higher reliability of the match between $p_a$ and $p_b$.}
    \vspace{-3mm}
    \label{fig:bidirectional}
\end{figure}

\subsection{Initial SfM via Dense Matching}

Our framework employs dense feature matching between image pairs, unlike traditional SfM pipelines~\cite{colmap} which rely on sparse keypoints~\cite{sift,superpoint}. By employing dense matching, our approach improves the reconstruction of texture-poor scenes and enables the registration of dense 3D points, which is challenging for conventional methods.

For efficiency, we first sample matches from the dense matching using non-max suppression~\cite{superpoint}, based on per-pixel confidence scores from the dense matching model’s output. Then, we sort out reliable matches by a method called mutual verification using bi-directional dense matching, similar to the mutual-nearest neighbor approach commonly used in detector-based methods.

As shown in Fig.~\ref{fig:bidirectional}, from the bi-directional dense matching results $M_{A\rightarrow{} B}$ and $M_{B\rightarrow{} A}$ between images $A$ and $B$, we define the pixel coordinates $p_b$ and $p_{a'}$ as
\begin{equation}
    \quad p_b = \mathbf{\mathcal{M}}_{A\rightarrow{} B}[p_a], \quad   p_{a'} =  \mathbf{\mathcal{M}}_{B\rightarrow{} A}[p_b]
\end{equation}    

where $p_{a'}$ is sampled using bilinear-interpolation.

To ensure more reliable matches, we filter out correspondences using a binary mask, which is obtained with the distance from $p_a$ to $p_{a'}$ as follows:
\begin{equation}
   B = [ || p_{a} - p_{a'} ||_2 \leq \epsilon_{\textit{p}}]
    \label{eq:bidirectional_distance}
\end{equation}

Using these filtered matches, we construct the initial SfM by triangulating 3D points through dense two-view fine matching. If camera poses are not provided for the reconstruction task, we first perform match quantization~\cite{detectorfreesfm} and incremental mapping~\cite{colmap} to obtain coarse poses. We then triangulate 3D points using raw matching with the coarse pose and refine both the point cloud and camera poses through the refinement process, as explained in Sec.~\ref{sec:sfm_refinement}.

\subsection{Track Extension via Gaussian Splatting}
\label{sec:track_extension_gs}
In the initial SfM, most of the tracks are short, \ie consisting of mostly two images, due to the fragmentary nature of feature tracks in pair-wise matching, which can limit the performance of track refinement, as shown in DFSfM~\cite{detectorfreesfm}.

Longer tracks, which offer more information on 3D geometry, could improve the track refinement performance. Thus, we generate new matching pairs on the existing matches for consistent multi-view matches, extending the length of track. We achieve this by projecting 3D points (triangulated by two-view matches) onto additional images on which, if successfully projected, the point is considered as visible from the camera pose as illustrated (by the blue dashed line) in Fig.~\ref{fig:overview}.

To determine the visibility of each 3D point from a camera pose, we apply a visibility filter using Gaussian Splatting (GS)~\cite{naive_gs, fsgs}, which supports fast training and rendering. We initialize a GS model with the 3D points obtained in the initial SfM. Specifically, we treat them as small 3D Gaussians setting each mean $\Mat \mu \in \mathbb{R}^3$ to the 3D point coordinates, rotation $R$ to the identity matrix, opacity $\alpha$ to 1, and scale $S$ based on the policy from Splatam~\cite{splatam}, which uses focal length and depth information. 
These Gaussians are optimized and densified to cover the entire scene, while keeping the initial Gaussian's parameters fixed. The details of initialization and training process is described in supplementary material.

After training the GS model, we utilize it for evaluating the visibility $M$ of all SfM points (represented as small Gaussians) from each camera pose using the following formula:
\vspace{-3mm}
\begin{equation}
\label{eq:render}
          M = \big[ \max_{r \in R}  \{ \alpha_{\text{SfM}}  \prod_{j=1}^{\text{$N_{SfM}$}-1}(1-\alpha_j) \} > \epsilon_{\textit{v}} \big]
\end{equation}

where $R$ is the set of rays from the camera center (of a camera pose) for rendering the image, $\alpha_{\text{SfM}}$ is the opacity of Gaussian representing the SfM point (set to $1$), $N_{SfM}$ the number of Gaussians along the ray $r$ from the camera center to the SfM point, and $\alpha_j$ the opacity of $j$-th intermediate Gaussian.
If the highest visibility score among the pixels rendered from the Gaussian representing the SfM point is larger than $\epsilon_{\textit{v}}$, we consider the SfM point, \ie $N_{SfM}$-th Gaussian, visible from the view of the camera pose. 
For each 3D point, we extend its track by adding its projected 2D point(s) onto the visible view(s). To do that, 
we project a 3D point $P_j$ onto the visible image using camera pose $\{R_i, t_i\}$ and intrinsic matrix $C_i$ as follows:
\begin{equation}
    p_{ij} =\Pi\left(\*R_i\*P_j+\*t_i, \*C_i\right) 
    \label{eq:fba}
\end{equation}


Then, we verify the correspondence between the newly obtained 2D point $p_{ij}$ and its associated original 2D keypoint with geometric verification~\cite{colmap} using initial SfM's camera poses. After verified, we add the new 2D point to the track of the 3D point $P_j$, which extends the length of $P_j$'s track.

Note that we project the 3D point onto the images that were not involved in the initial triangulation (\ie, establishing a new one-to-one correspondence of 2D-3D matches per image as illustrated with green lines in Fig.~\ref{fig:overview} (Stage 2). 
As will be demonstrated in our experiments (Tab.~\ref{tab:ablation}), the proposed \textit{Track extension} enables the subsequent track refinement to produce more refined tracks, which leads to more accurate 3D points in the SfM model.

\subsection{Iterative SfM Refinement}
\label{sec:sfm_refinement}
Using the extended tracks, we refine the initial SfM model to obtain improved camera poses and point cloud. Specifically, we iteratively improve the accuracy of feature tracks by employing our multi-view kernelized matching module alongside bundle adjustment, jointly optimizing camera poses and point cloud.

\subsubsection{Feature Track Refinement}

\begin{figure}
    \centering
    \resizebox{\linewidth}{!}{
    
    \raisebox{-0.5\height}
    {\includegraphics{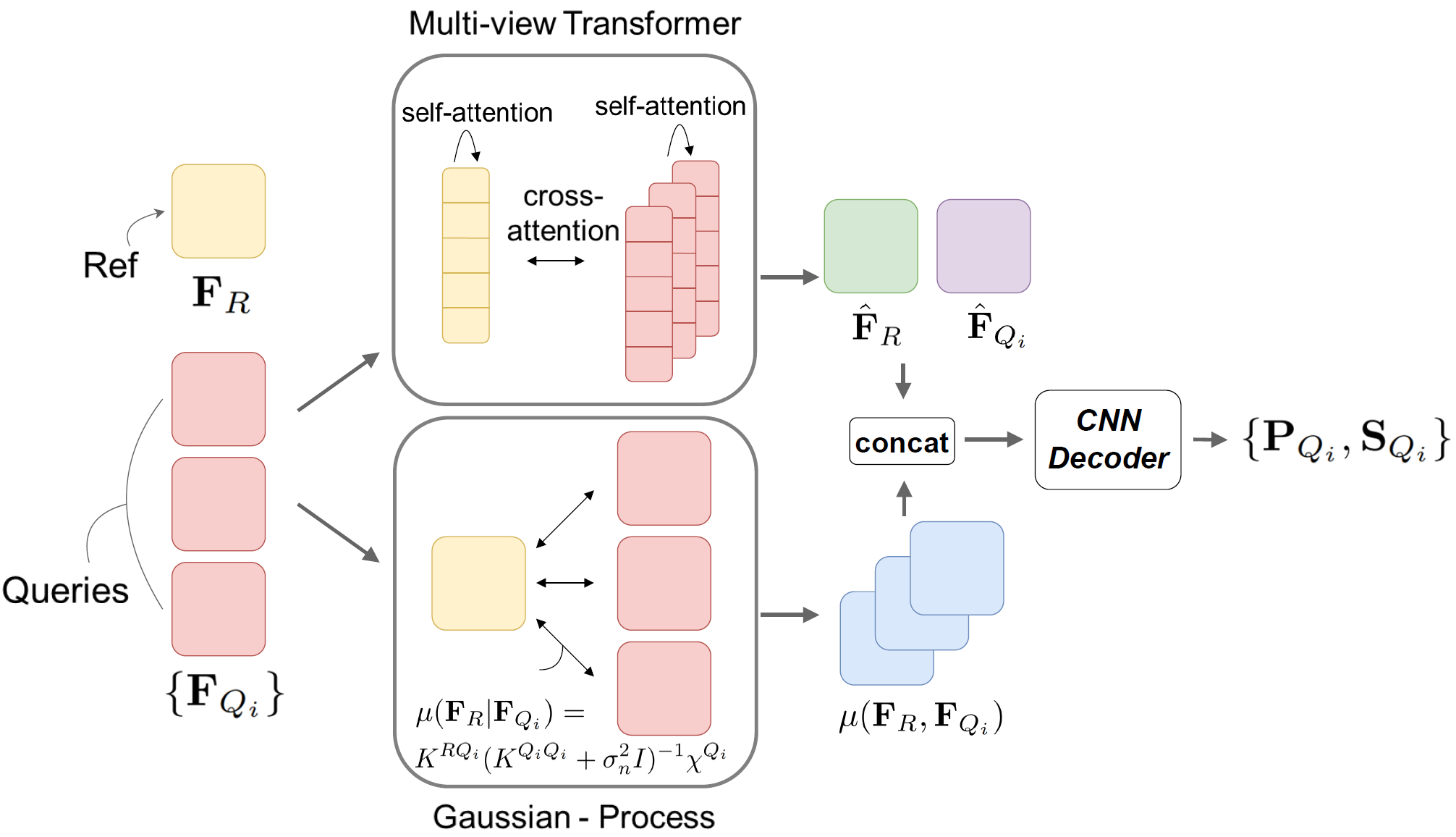}}}
    \caption{\textbf{Multi-view kernelized matching architecture.}
The feature maps from the reference and query views are processed in two ways: first, by a multi-view transformer, applying self and cross-attention and secondly, by a Gaussian Process that computes the posterior mean conditioned on the feature of each query view. All results are concatenated and fed into a CNN decoder, yielding probability distribution of coordinates for refinement $\mathbf{P}_{Q_i}$ and corresponding confidence score $\mathbf{S}_{Q_i}$ for each query view. }

\vspace{-0.4cm}
    \label{fig:trackrefine}
\end{figure}

\begin{figure}
    \centering
    \resizebox{\linewidth}{!}{
    
    \raisebox{-0.5\height}
    {\includegraphics{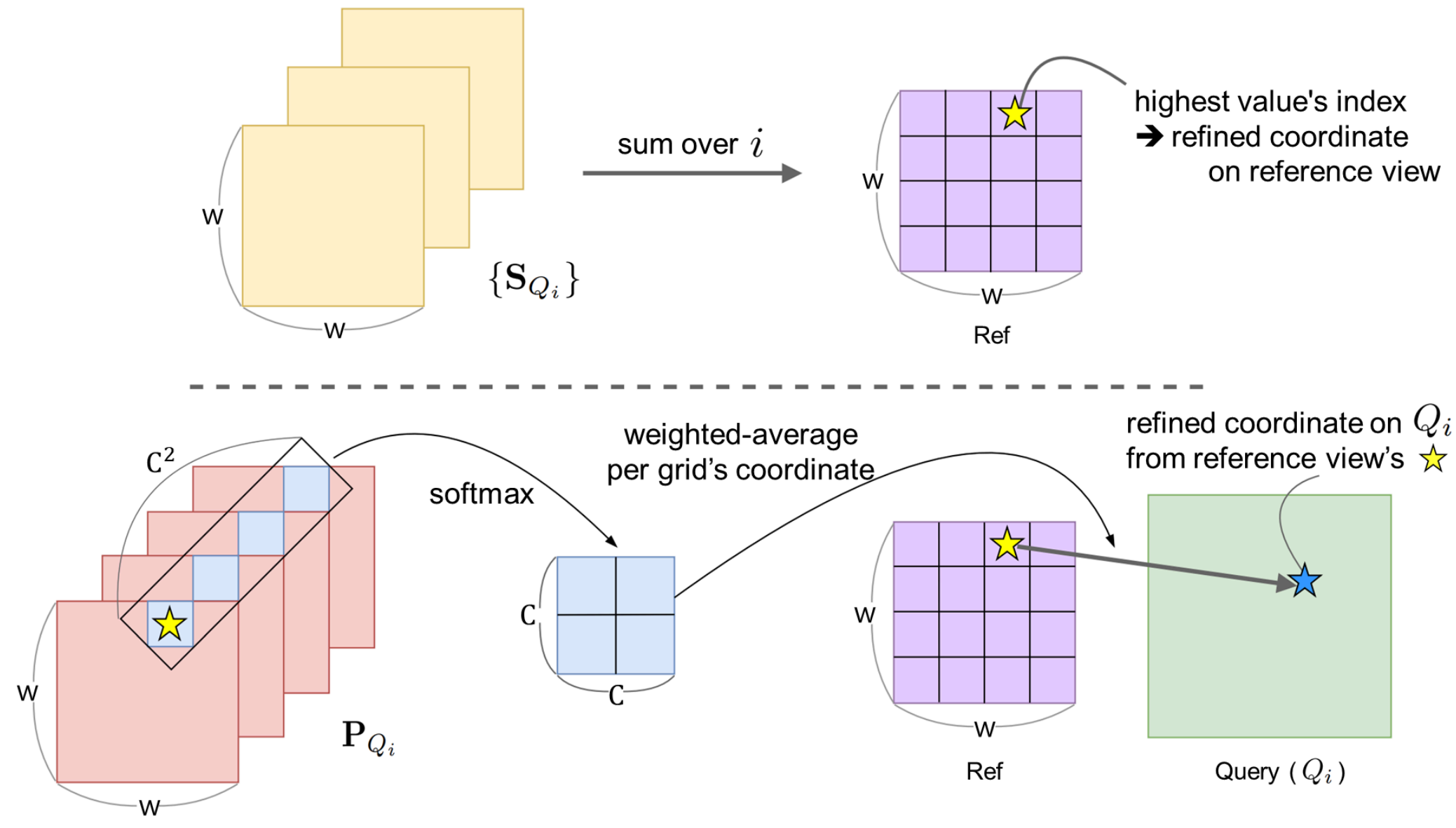}}}
    \vspace{-2mm}
    \caption{\textbf{Regression of refined coordinate.}
     \textbf{Top: Selection of refined coordinate on the reference view:} From the set of $S_{Q_i}$ values, representing the correspondence scores from the reference view to each query view $Q_i$, we sum over $S_{Q_i}$'s and select the pixel with the highest score as the refined coordinate for the reference view.
    \textbf{Bottom: Regression of refined coordinate on each query view:} We apply a channel-wise softmax to obtain a probability distribution of coordinates for each pixel, then calculate a weighted average to determine the correspondence from the reference view to query view $Q_i$.
   }
    \vspace{-0.5cm}
    \label{fig:regress}
\end{figure}

A feature track (or track) $\mathcal{T}_j = \{\mathbf{x}_k \in \mathbb{R}^2 | k=1:N_j\}$ represents a set of 2D keypoint locations on multi-view ($N_j$) images that correspond to a 3D point $P_j$. 
In our proposed method of feature track refinement, each track goes through a multi-view kernelized matching architecture (Fig.~\ref{fig:trackrefine}) and the regression of refined coordinates (Fig.~\ref{fig:regress}). The matching architecture produces two outputs, a probability distribution of coordinate per pixel $P_{Q_i}$ and confidence score $S_{Q_i}$ per view. Then, the regression from these outputs produces a refined track by adjusting the keypoint locations.

\paragraph{Multi-view kernelized matching.}
Following DFSfM, we split a track into the keypoint of a reference view and those of query views (views from the track except the reference view) with a policy of median depth value. 
Then, we feed image patches centered at each keypoint into a CNN feature extractor, obtaining feature maps ${\mathbf{F}_R, \mathbf{F}_{Q_i} \in \mathbb{R}^{p \times p \times c} }$, where $\mathbf{F}_R$ is from the reference view and $\mathbf{F}_{Q_i}$'s are from query views.
As shown in Fig.~\ref{fig:trackrefine}, our multi-view kernelized matching architecture takes as input these feature maps of reference and query views.

Our proposed matching architecture consists of two paths, feature-based and coordinate-embedding-based ones.
On the feature-based path (denoted with Multi-view Transformer in Fig.~\ref{fig:trackrefine}), we utilize a local feature transformer to obtain the transformed feature map $\hat{\mathbf{F}}_R$ and $\hat{\mathbf{F}}_{Q_i}$'s through self and cross-attention between $\mathbf{F}_R$ and $\mathbf{F}_Q$'s. 
On the coordinate-embedding-based path (denoted with Gaussian-Process in Fig.~\ref{fig:trackrefine}), we employ a Gaussian Process~\cite{gaussian_process} to utilize the positional information, deriving a coordinate embedding feature $\mu(\mathbf{F}_R | \mathbf{F}_{Q_i})$ for each query view. Following DKM~\cite{dkm}, we use an exponential cosine similarity kernel:

\vspace{-5mm}
\begin{equation}
    k(f_A,f_B) = \exp(-\tau)\exp\bigg(\tau\frac{\langle f_A, f_B \rangle}{\sqrt{\langle f_A, f_A  \rangle \langle  f_B, f_B  \rangle+\varepsilon}}\bigg),
\end{equation}

where $f_A$ and $f_B$ each is a feature vector from $\mathbf{F}_A$ and $\mathbf{F}_B$, respectively.
From this kernel function, we build kernel matrices $ K^{\R\Q_i}, K^{\Q_i\Q_i}  \in \mathbb{R}^{p^2 \times p^2}$ to obtain posterior mean $\mu(\mathbf{F}_R | \mathbf{F}_{Q_i})$ as follows: 
\vspace{-1mm}
\begin{equation}
    \mu(\mathbf{F}_R | \mathbf{F}_{Q_i}) = K^{R Q_i}(K^{Q_i Q_i}+\sigma_n^2 I)^{-1}{\chi}^{Q_i}
\end{equation}

where $\chi^{\Q_i}$ is the Fourier positional encoding~\cite{nerf} of $F_{Q_i}$'s pixel coordinate. Finally, we concatenate the transformed feature maps $\hat{\mathbf{F}}_R$,  $\hat{\mathbf{F}}_{Q_i}$ and coordinate embedding $\mu(\mathbf{F}_R | \mathbf{F}_{Q_i})$, and feed them into a CNN decoder $D_{\theta}$:

\vspace{-3mm}
\begin{equation}
   P_{Q_i}, S_{Q_i} =  D_{\theta}( \hat{\mathbf{F}}_R \oplus  \hat{\mathbf{F}}_{Q_i} \oplus \mu(\mathbf{F}_R | \mathbf{F}_{Q_i}))
\end{equation} 

where we obtain a probability distribution of coordinate per pixel $P_{Q_i} \in \mathbb{R}^{C^2 \times w \times w}$ and confidence score $S_{Q_i} \in \mathbb{R}^{w \times w}$, $C^2$ is the number of distribution's anchors, and $w \times w$ a search boundary for refined coordinates.

\paragraph{Regression of keypoint's location.}

Fig.~\ref{fig:regress} illustrates how the refined track is obtained.
First, we obtain the refined coordinate of keypoint on the reference view.
From the output of multi-view kernelized matching, $\{S_{Q_i}\}$, we estimate the confidence value of each pixel as the candidate of refined coordinate on the reference view. 
As shown in Fig.~\ref{fig:regress} (top), we select, as the refined coordinate on the reference view, the pixel with the highest value, \ie the highest confidence from $\sum_i S_{Q_i}$. 

Then, as shown in Fig.~\ref{fig:regress} (bottom), from the refined coordinate on the reference view (calculated above), we estimate the refined coordinate of keypoint for each query view $Q_i$. For each patch of $P_{Q_i}$ of size $w \times w$, we apply softmax across channels from the pixel of refined coordinate for the reference view, to obtain a probability distribution over a $C \times C$ grid. The refined coordinate for the query view $Q_i$ is then computed as a weighted average, where each pixel’s probability serves as its weight. By multiplying each probability by its corresponding pixel coordinate and summing these weighted values, we derive a refined coordinate for query view $Q_i$. Finally, we construct the final refined track ${\mathcal{T}^*_j}$, consisting of the refined coordinates for the reference view and all query views.

\paragraph{Training setup and loss function.}
We set up training as in DFSfM~\cite{detectorfreesfm} using MegaDepth dataset~\cite{megadepth}. We use image bags sampled from the SfM model provided in the dataset,
in which ground truth tracks are derived from the given camera pose and MVS depth map. For training, we perturb the ground truth 2D locations of tracks in the query views with random noise, providing these as input to our refinement module.
Unlike DFSfM, which relies on statistics (variance) as uncertainty measure, we directly learn confidence score through a neural network. Consequently, our loss function is not simply formulated as an $L_2$ loss to minimize the distance between the refined tracks and the ground truth tracks. Instead, we jointly predict and learn the confidence $ S_{Q_i} > 0$ for each query view, and include it in the regression loss as follows:
\begin{equation}
    \mathcal{L} = \frac{1}{N} \sum_{j \in n_t} \sum_{i \in n_j}  s_{Q_i} \cdot \left\lVert p_{Q_i} - p_{gt} \right\rVert_2 - \alpha \log  s_{Q_i}.
    \label{eq:loss}
\end{equation}
where $n_t$ is the number of feature tracks, $n_j$ the number of query views in a track $j$, $N$ the total number of refined keypoints on all the query views, $p_{Q_i}$ and $p_{gt}$ the learned 2D coordinate from $Q_i$ (as illustrated in Fig.~\ref{fig:regress}) and the ground truth 2D coordinate, respectively, 
and $s_{Q_i}$ the confidence score of prediction $p_{Q_i}$, which is obtained, from $S_{Q_i}$, at the pixel of the reference view's ground truth keypoint.
The second term in the loss is used to avoid being under-confident, where we set $\alpha$ as 20.

\subsubsection{Bundle Adjustment}

Using the refined feature tracks, we iteratively perform geometric bundle adjustment (BA)~\cite{bundleadjustment} to refine the intrinsic parameters~$\{\*C_i\}$, camera poses~$\{\boldsymbol{\xi}_i\}$, and 3D points' coordinates~$\{\*P_j\}$. Formally, we minimize the reprojection error $E$ defined as: 
\begin{equation*}    
    \resizebox{0.8\linewidth}{!}{
    $
    E = \sum_j \sum_{\*x_k^* \in \mathcal{T}^*_j} \rho \bigl( \left\lVert \pi \left(\boldsymbol{\xi}_i \cdot \*P_j, \*C_i\right) - \*x^*_k \right\rVert_2^2 \bigr) \enspace,
    $
    }
    \label{eq:gba}
\end{equation*}
where $\pi(\cdot)$ projects points to image plane using intrinsic parameter $\*C_i$, and $\rho(\cdot)$ is a robust loss function~\cite{hampel1986robust}. We then filter outliers~\cite{incrementalsfm, colmap} by rejecting points that exceed the maximum reprojection error threshold $\epsilon_f$. After performing BA and outlier filtering, the refined SfM can be passed to the next iteration of feature track refinement and BA.

\section{Experiments}

\subsection{Datasets}
We use the ETH3D~\cite{eth3d_dataset}, Texture-Poor SfM datasets~\cite{detectorfreesfm}, and Image Matching Challenge (IMC) 2021 dataset~\cite{imc_dataset}, each presenting unique challenges across different types of scenes. The ETH3D dataset includes indoor and outdoor scenes with sparsely captured, high-resolution images, and ground truth poses calibrated using Lidar. The Texture-Poor SfM dataset focuses on object-centric scenes with low texture, comprising subsampled image sets. 
We use all test scenes from the IMC Phototourism dataset, evaluating them with subsampled image bags. The main challenge of this dataset lies in its sparse views, accompanied by significant variations in viewpoint and illumination.

\subsection{Implementation Details}

For efficiency, we apply non-max suppression~\cite{superpoint} with a radius of 4 pixels for ETH3D and 2 pixels for Texture-Poor SfM and IMC dataset.
For mutual verification, we filter out matches with a distance exceeding $\epsilon_{\textit{p}} = 3$px in Eq.~\ref{eq:bidirectional_distance}.

To determine the visibility of Gaussians (SfM points) from a camera pose, we set the visibility threshold $\epsilon_{\textit{v}}$ to 0.5 in Eq.~\ref{eq:render}. Details of training and implementation of Gaussian Splatting are provided in the supplementary material.

We use S2Dnet~\cite{s2dnet} as a CNN backbone for our refinement architecture, and adopt the multi-view transformer architecture from DFSfM~\cite{detectorfreesfm}. To ensure a fair comparison, we apply the same refinement parameters as in DFSfM, including a local patch size ($p=15$) from the feature maps ($\mathbf{F}_R$ and $\mathbf{F}_Q$) obtained from the CNN backbone, and a search boundary for reference view's refined keypoints set to $w=7$. The number of distribution anchors $C^2$ is set to $7^2$. For the CNN decoder in Fig.~\ref{fig:trackrefine}, we use a series of residual blocks~\cite{resnet} and channel attention blocks~\cite{cab}. During bundle adjustment, matches with reprojection errors exceeding $\epsilon_{\textit{f}} = 3$px are discarded. The total SfM refinement process is iterated twice. More implementation details are reported in the supplementary material.

\subsection{3D Triangulation}
Accurately triangulating scene point clouds, given known camera poses and intrinsics, is an important task in SfM. In this section, we evaluate the accuracy and completeness of the reconstructed point clouds.

\paragraph{Evaluation.}
We use the ETH3D training set, comprising 13 indoor and outdoor scenes with millimeter-accurate dense point clouds as ground truth. Following the protocol in PixSfM~\cite{pixsfm} and DFSfM~\cite{detectorfreesfm}, we triangulate 3D points using fixed camera poses and intrinsics, then evaluate the resulting point clouds for accuracy and completeness using the ETH3D benchmark~\cite{eth3d_dataset}. The results are reported at varying distance thresholds (1cm, 2cm, 5cm), averaged across all scenes. Consistent with the benchmark setup, accuracy and completeness are measured by the proportion of triangulated points that fall within the specified distances from the ground truth points.

\begin{table}[t]
    \centering
    \resizebox{1.0\columnwidth}{!}{
    \setlength\tabcolsep{6pt} %
    \begin{tabular}{cccccccc} 
    \toprule
    \multirow{2}{*}{Type}  &\multirow{2}{*}{Method}         & \multicolumn{3}{c}{Accuracy~($\%$)}             & \multicolumn{3}{c}{Completeness~($\%$)} \\ 
    \cmidrule(lr){3-5}
    \cmidrule(lr){6-8}
                            & & 1cm       & 2cm       & 5cm       & 1cm       & 2cm       &  5cm \\ 
    \midrule
    \multirow{4}{*}{\begin{tabular}[c]{@{}c@{}}Detector-\\ Based\end{tabular}} & SIFT~\cite{sift} + NN + PixSfM~\cite{pixsfm} & 76.18 & 85.60 & 93.16 & 0.17 & 0.71 & 3.29\\
    &D2Net~\cite{d2net} + NN + PixSfM & 74.75 & 83.81 & 91.98 & 0.83 & 2.69 & 8.95\\
    &R2D2~\cite{r2d2} + NN + PixSfM & 74.12 & 84.49 & 91.98 & 0.43 & 1.58 & 6.71\\
    &SP~\cite{superpoint} + SG~\cite{superglue} + PixSfM & 79.01 & 87.04 & 93.80 & 0.75 & 2.77 & 11.28 \\
    \midrule
    \multirow{4}{*}{\begin{tabular}[c]{@{}c@{}}Semi-\\ Dense \\ Matching \end{tabular}}     
    & LoFTR~\cite{loftr} + PixSfM & 74.42 & 84.08 & 92.63 & 2.91 & 9.39 & 27.31 \\
    & LoFTR + DFSfM~\cite{detectorfreesfm} &  80.38 & 89.01 & 95.83 & 3.73 & 11.07 & 29.54 \\ 
    & AspanTrans.~\cite{Chen2022ASpanFormerDI} + DFSfM & 77.63 & 87.40 & 95.02 & 3.97 & 12.18 & 32.42\\ 
    & MatchFormer~\cite{wang2022matchformer} + DFSfM & 79.86 & 88.51 & 95.48 & 3.76 & 11.06 & 29.05\\
    \midrule
    \multirow{3}{*}{\begin{tabular}[c]{@{}c@{}}  Dense- \\ Matching \end{tabular}}     
    & RoMa + DFSfM & 79.32 & 88.42 & 95.82 & 3.13 & 9.79 & 29.10 \\ 
    & DKM + Ours &  \underline{84.05} & \underline{91.96} & \underline{97.39} & {5.95} & {13.95} & {31.03} \\  
    & RoMa + Ours &  \textbf{84.79} & \textbf{92.62} & \textbf{97.77} & \underline{7.38} & \underline{17.06} & \underline{36.35} \\ 

    \midrule
    \multirow{1}{*}{Deep} 
    & VGGSfM~\cite{vggsfm} &  80.62 & 89.49 & 96.52 & 4.52 & 13.11 & 33.96 \\
    \midrule
    \multirow{3}{*}{Point-Based} 
    & Mast3r~\cite{mast3r} + COLMAP &  42.02 & 59.74 & 80.58 & 2.05 & 8.76 & 31.73\\    
    & Mast3r + COLMAP + Our Refinement (Sec.~\ref{sec:sfm_refinement}) &  77.22 & 86.47 & 94.05 & 2.40 & 8.26 & 27.17 \\
    & Mast3r-SfM\tablefootnote{The Mast3r-SfM results have been updated as of January 2026.} ~\cite{mast3r-sfm} &  27.34 & 43.90 & 69.50 & \textbf{14.86} & \textbf{34.81} & \textbf{68.72}\\
    
    \bottomrule
    \end{tabular}
    }

    \vspace{-0.1cm}
    \caption{\textbf{Results of 3D Triangulation.}
    Our method is compared with the baselines on the ETH3D~\cite{eth3d_dataset} dataset using accuracy and completeness metrics with different thresholds.
    }
    \label{tab:exptriangulation}
    \vspace{-0.6cm}
\end{table}

\paragraph{Results.}

The results are shown in Tab.~\ref{tab:exptriangulation}. While there is a trade-off between accuracy and completeness, our Dense-SfM framework achieves superior performances on both metrics. Compared to the state-of-the-art baseline LoFTR + DFSfM, our pipeline with RoMa achieves higher accuracy and completeness, thanks to employing dense matching and our proposed pipeline. Also, compared to the RoMa + DFSfM, where we apply quantization on dense matching and refine the tracks as in DFSfM, our method achieves higher performance in terms of both accuracy and completeness, highlighting the effectiveness of our SfM framework. More results with detector-based and semi-dense matching are reported in the supplementary material.

\begin{table*}[ht!]
    \centering
    \resizebox{0.95\textwidth}{!}{
    \setlength\tabcolsep{6pt} %
    \begin{tabular}{ccccccccccc} 
    \toprule
    \multirow{2}{*}{Type} & \multirow{2}{*}{Method}         & \multicolumn{3}{c}{ETH3D Dataset}             &  \multicolumn{3}{c}{Texture-Poor SfM Dataset}  & 
    \multicolumn{3}{c}{IMC Dataset}
    \\ 
    \cmidrule(lr){3-5}
    \cmidrule(lr){6-8} \cmidrule(lr){9-11} &     & AUC@1$\degree$       & AUC@3$\degree$       & AUC@5$\degree$  & AUC@3$\degree$ & AUC@5$\degree$ & AUC@10$\degree$ & AUC@3$\degree$ & AUC@5$\degree$ & AUC@10$\degree$ \\ 
    \midrule
    \multirow{5}{*}{Detector-Based} &   COLMAP~(SIFT+NN)~\cite{colmap} & 26.71 & 38.86 & 42.14 &  2.87 & 3.85 & 4.95 & 23.58 & 23.66 & 44.79 \\
    
    &   SIFT~\cite{sift} + NN + PixSfM~\cite{pixsfm} & 26.94 & 39.01 & 42.19  & 3.13 & 4.08 & 5.09 & 25.54 & 34.80 & 46.73 \\

    &   D2Net~\cite{d2net} + NN + PixSfM & 34.50 & 49.77 & 53.58 & 1.54 & 2.63 & 4.54 & 8.91 & 12.26 & 16.79 \\
    
    &   R2D2~\cite{r2d2} + NN + PixSfM & 43.58 & 62.09 & 66.89 & 3.79 & 5.51 & 7.84 & 31.41 & 41.80 & 54.65 \\
    &   SP~\cite{superpoint} + SG~\cite{superglue} + PixSfM & 50.82 & 68.52 & 72.86 & 14.00 & 19.23 & 24.55 & 45.19 & 57.22 & 70.47 \\
    
    \hline
    
    \multirow{4}{*}{Semi-Dense Matching} 
    & LoFTR~\cite{loftr} + PixSfM & 54.35 & 73.97 & 78.86 & 20.66 & 30.49 & 42.01 & 44.06 & 56.16 & 69.61  \\
    & LoFTR + DFSfM~\cite{detectorfreesfm} & \underline{59.12} & 75.59 & 79.53 & 26.07 & 35.77 & 45.43 & 46.55 & 58.74 & 72.19  \\
    & AspanTrans.~\cite{Chen2022ASpanFormerDI} + DFSfM & 57.23 & 73.71 & 77.70 & 25.78 & 35.69 & 45.64 & 46.79 & 59.01 & 72.50 \\
    & MatchFormer~\cite{wang2022matchformer} + DFSfM & 56.70 & 73.00 & 76.84  & 26.90 & 37.57 & 48.55 & 45.83 & 57.88 & 71.22 \\

    \hline

    \multirow{3}{*}{Dense Matching} 
    & RoMa + DFSfM & 57.97 & 77.02& 81.59 & 34.28 & 50.01 & 69.01 & 47.43& 59.84 & 73.19 \\
    & DKM + Ours & 59.04 &  \underline{77.73} &  \underline{82.20} & 41.19& 57.26 & 72.95 & \textbf{48.65} & \textbf{61.09} & \textbf{74.37}  \\
    & RoMa + Ours & \textbf{60.92} & \textbf{78.41} &  \textbf{82.63} & \textbf{49.94} & \textbf{66.23} & \textbf{81.41} & \underline{48.48} & \underline{60.79} & 73.90 \\

    \hline
    \multirow{1}{*}{Deep} 
    & VGGSfM~\cite{vggsfm} & 28.09 & 44.41 & 49.91 & 1.86 & 6.26 & 21.10 & 45.23 & 58.89 & \underline{73.92}  \\

    \hline
    \multirow{3}{*}{Point-Based} 
    & Mast3r\cite{mast3r} + COLMAP & 35.85 & 58.46 & 65.03 & 43.46  & 61.51 & 78.97 & 42.26 & 54.53 & 67.97 \\
    & Mast3r + COLMAP + Our Refinement & 37.50 & 59.18 & 65.48 & \underline{48.03} & \underline{64.48} & \underline{80.31} & 44.98 & 57.09 & 70.09 \\
    & Mast3r-SfM\tablefootnote{The Mast3r-SfM results have been updated as of January 2026.} ~\cite{mast3r-sfm} & 27.05 & 45.84 & 52.61 & 7.10 & 18.08 & 42.44 & 31.77 & 46.36 & 64.37 \\

    \bottomrule
    \end{tabular}
    }
    \vspace{-0.1cm}
    \caption{\textbf{Results of Multi-View Camera Pose Estimation.}
    Our framework is compared with detector-based, detector-free and dense-matching baselines by the AUC of pose error at different thresholds.
    \textbf{Bold} and \underline{underline} indicate the best and second-best results.
    }
    \label{tab:exp_camerapose}
    \vspace{-0.35 cm}
    \end{table*}

\subsection{Multi-View Camera Pose Estimation}
In this section, we evaluate the recovered multi-view poses.

\paragraph{Evaluation protocols.}
For the evaluation, we use a subset of ETH3D training and test dataset, which includes 22 indoor and outdoor scenes, the Texture-Poor SfM dataset, and the test scenes of IMC 2021 dataset. We evaluate the accuracy of the estimated multi-view poses using the AUC of pose error at various thresholds. For the evaluation of IMC 2021 dataset, due to the high occlusion and appearance changes in many test scenes, directly applying track extension via GS is challenging. Instead, we adopt a quantization method~\cite{detectorfreesfm} on the matching results, focusing on comparing our multi-view kernelized matching module with DFSfM's refinement module. Specifically, we apply quantization to the matched keypoints with a radius of $4$, and then use our refinement module to refine the quantized keypoints.

\begin{figure}[h]
    \vspace{-0.3cm}
    \centering
    \resizebox{0.95\linewidth}{!}{
    
    \raisebox{-0.5\height}
    {\includegraphics{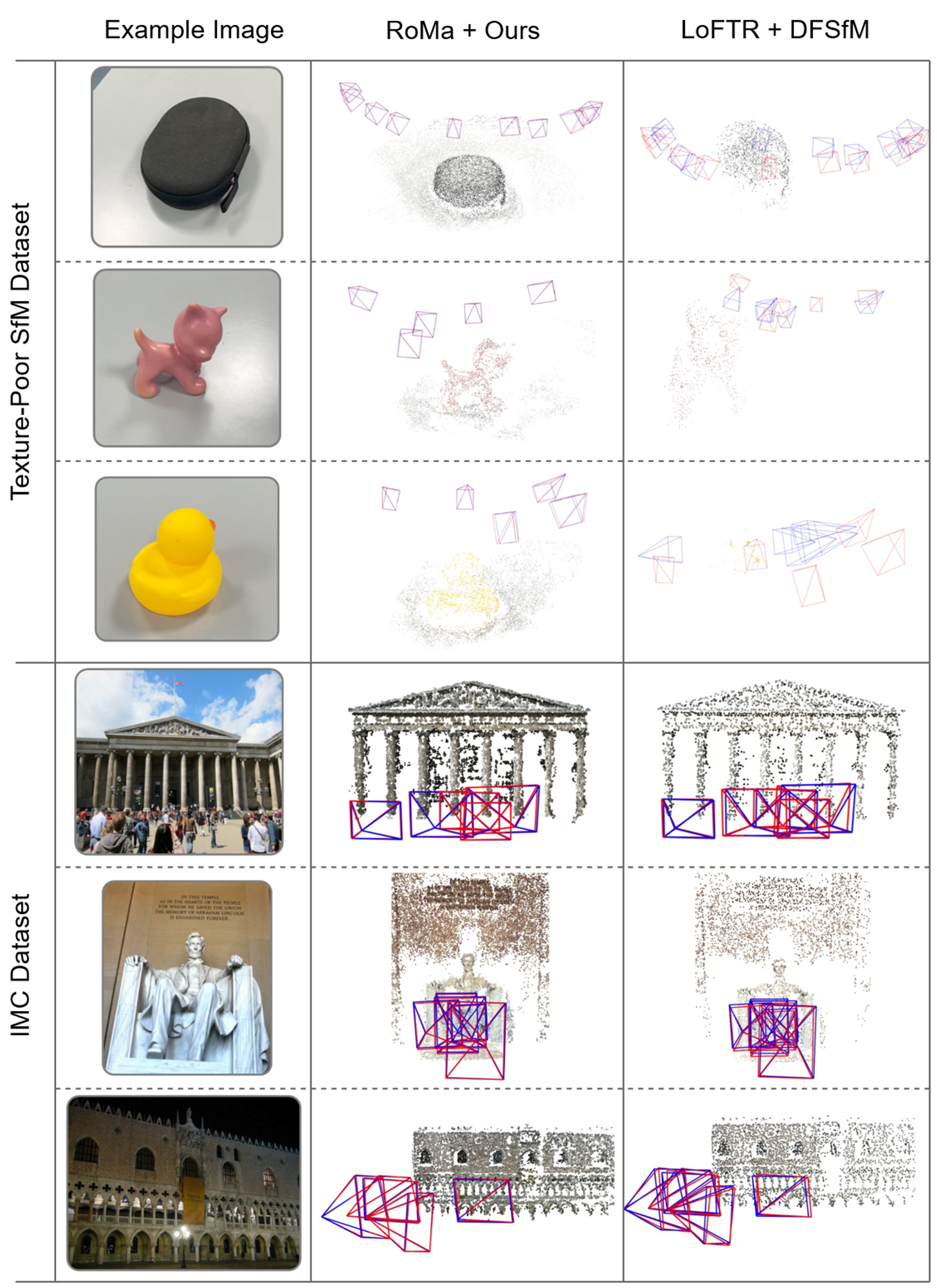}}}
    \vspace{-2mm}
    \caption{\textbf{Qualitative Results on Texture-Poor SfM and IMC dataset.} Our method with the dense matcher RoMa is qualitatively compared with the baseline LoFTR + DFSfM. The red cameras are ground truth poses while the blue cameras are recovered poses.}
    \vspace{-0.8cm}
    \label{fig:visualize}
\end{figure}

\paragraph{Results.}
As shown in Tab.~\ref{tab:exp_camerapose}, our Dense-SfM framework outperforms existing baselines over all datasets. On the ETH3D dataset, our framework with dense matching achieves the highest multi-view pose accuracy, with a particularly large improvement at the highest threshold.

As shown in Fig.~\ref{fig:visualize}, our framework successfully reconstructs dense and accurate 3D points thanks to dense matching and our pipeline, even in challenging low-textured regions with viewpoint changes, leading to much higher pose estimation accuracy.

\subsection{Ablation Studies}

We conduct several experiments to evaluate the key components of our pipeline, including the matching design used as input for SfM and the architectural choices of the refinement module, on the ETH3D dataset using triangulation metrics.

\begin{table}[ht!]
    \centering
    \resizebox{1.0\columnwidth}{!}{
    \setlength\tabcolsep{6pt} %
    \begin{tabular}{cccccc}
    \toprule
    \multirow{2}{*}{}  & \multirow{2}{*}{Design Choice}         & \multicolumn{2}{c}{Accu.~($\%$)}             & \multicolumn{2}{c}{Complete.~($\%$)}\\ 
    \cmidrule(lr){3-4}
    \cmidrule(lr){5-6}
        &                    & 1cm       & 2cm       & 1cm       & 2cm  \\ 
    \midrule

     \multirow{4}{*}{(1) Matching designs}&
     \textbf{Full model} & 84.79 & 92.62 & 7.38 & 17.06  \\
    & w/o Mutual Verification & 84.03 & 92.06 & 8.14 & 18.68\\
    &  w/o GS Extension & 77.45 & 87.80 & \textbf{9.45} & \textbf{21.64} \\
    & Quantized Matching (r=4) & 82.82 & 90.75 & 3.53 & 10.62 \\

    \midrule

    \multirow{5}{*}{(2) Refinement designs} 
    & \textbf{Full model} & 84.79 & 92.62 & 7.38 & 17.06  \\
    & Multi-view Trans.~\cite{detectorfreesfm} & 80.59 & 90.04 & 6.04 & 15.2 \\
    
    & w/o Gaussian Process & 72.47 & 84.72 & 4.78 & 12.59 \\ 
    & w/o Multi-view Trans. & 83.39 & 91.63 & 7.04 & 16.49  \\ 
    & w/o ref. confidence selection & 83.81 & 92.15 & 8.83 & 19.85\\
    \midrule    

    \multirow{4}{*}{(3) Number of iterations} & No refine. & 69.77 & 82.59 & 8.61 & 20.4 \\
    & 1 iter & 81.12 & 90.52 & 7.62 & 17.86 \\
    & 2 iter & 84.79 & 92.6 & 7.38 & 17.06   \\
    & 3 iter & \textbf{85.01} & \textbf{92.75} & 6.74 & 15.91\\

    \bottomrule
    \end{tabular}
    }

    \vspace{-0.2cm}
    \caption{\textbf{Ablation Studies.}
On the ETH3D dataset, we quantitatively evaluate the impact of design choice of matching and refinement module, and the number of iterations of refinement. The reported triangulation accuracy and completeness are averaged across all scenes.    
    }
    \label{tab:ablation}
    \vspace{-0.6 cm}
\end{table}

\paragraph{Matching Designs.}
Tab.~\ref{tab:ablation} (1) demonstrates the benefits of our proposed matching filtering (mutual verification) and track extension via Gaussian Splatting (GS). Compared to the configurations without mutual verification, where all matches are used for reconstructing the initial SfM, incorporating mutual verification improves accuracy, with a trade-off in completeness due to filtering. 
Moreover, extending match pairs through GS significantly improves accuracy, which shows that using more views for multi-view kernelized matching contribute to higher performance of accuracy.

For a comparison with DFSfM~\cite{detectorfreesfm}, which applies quantization to matched keypoints to address the fragmented track problem, we replaced our GS track extension with their quantization approach, using a rounding ratio of $r=4$. Results indicate that our GS track extension not only maintains completeness by avoiding the limitations of quantization but also achieves superior accuracy.

\paragraph{Refinement designs.}
Table~\ref{tab:ablation} (2) shows the benefit of our proposed multi-view kernelized matching's components in Fig.~\ref{fig:trackrefine} and confidence-based track selection in Fig.~\ref{fig:regress}.

We first compare our refinement module to the multi-view refinement module in DFSfM~\cite{detectorfreesfm}, by substituting our refinement module with it. 
The result shows that our refinement module consistently outperforms the existing approach, achieving higher accuracy and completeness.

Next, we evaluate the contributions of each component in our refinement architecture (Fig.~\ref{fig:trackrefine}) by testing configurations where either the multi-view transformer or Gaussian Process is removed. The results show that our full model, which integrates both components, achieves the highest performance in accuracy and completeness, confirming that combining feature-based and positional information yields more precise refinement.
Furthermore, our strategy of selecting a final refined track based on predicted confidence in the reference view further enhances accuracy.

\paragraph{Impact of iterations.}
Table~\ref{tab:ablation} (3) reports the results after each refinement iteration. The initial SfM point cloud is less accurate due to reliance solely on initial two-view matching. As the iterations progress, the accuracy of the SfM point cloud gets improved. However, further refinement beyond two iterations gives little gains. So, we limit the refinement to two iterations for efficiency.

\section{Conclusion}
This paper introduces a novel Dense-SfM framework aimed at reconstructing dense point cloud and estimating camera poses from a set of images. Unlike conventional SfM systems that often rely on initial 2D matches, our framework leverages Gaussian Splatting to generate new matching pairs, producing longer tracks that significantly enhance both the accuracy and density of the reconstructed point cloud. Extensive evaluations on diverse datasets show that our framework consistently outperforms existing SfM baselines across multiple metrics.

\section{Acknowledgement}
This research was supported in part by Samsung Advanced Institute of Technology (SAIT), the NAVER-Intel Co-Lab, Inter-university Semiconductor Research Center (ISRC), and Institute of Information \& communications Technology Planning \& Evaluation (IITP) grant funded by the Korea government (MSIT) [NO.RS-2021-II211343, Artificial Intelligence Graduate School Program (Seoul National University)].

{
    \small
    \bibliographystyle{ieeenat_fullname}
    \bibliography{main}
}
\clearpage
\setcounter{page}{1}
\maketitlesupplementary

\section{Experiment}

\subsection{Sparse feature matching with our refinement}

\begin{table}[ht!]
    \centering
    \resizebox{1.0\columnwidth}{!}{
    \setlength\tabcolsep{3pt} %
    \begin{tabular}{c| c|ccc|ccc}
    \toprule
    \multirow{2}{*}{Sparse Det. \& Matcher} & \multirow{2}{*}{Refinement}  & \multicolumn{3}{c}{Accuracy~($\%$)} & \multicolumn{3}{c}{Completeness~($\%$)} \\
    \cmidrule(lr){3-5}
    \cmidrule(lr){6-8}
    & & 1cm       & 2cm       & 5cm       & 1cm       & 2cm       &  5cm \\ 
    \midrule
    \multirow{4}{*}{SIFT + NN} 
    & Raw & 66.4 &  78.62 &  89.38 & 0.12 & 0.60 &  3.02 \\
    & PixSfM & 76.81& 86.72 & 94.28 & \textbf{0.15} & \textbf{0.66} &  \textbf{3.12} \\
    & DFSfM & \textbf{82.66} & 90.41 & \textbf{96.3} & 0.14 & 0.59 & 2.78\\
    & Ours & 82.37 & \textbf{90.42} & \textbf{96.3} & \textbf{0.15} & 0.63 & 3.01 \\
    
    \midrule
    
    \multirow{4}{*}{R2D2 + NN} 
    & Raw & 59.7 & 73.92  & 87.25 & 0.34 & 1.4 & 6.42 \\    
    & PixSfM  & 76.31 & 85.62 & 93.37 & 0.42 & 1.52 & 6.43 \\
    & DFSfM &  81.51 & 89.81 & 95.81 & \textbf{0.47} & \textbf{1.63} & \textbf{6.57}  \\
    & Ours & \textbf{83.1} & \textbf{90.5} & \textbf{96.03} & 0.45 & 1.58 & 6.54 \\
    \midrule
    
    \multirow{4}{*}{SP + SG} 
    & Raw & 67.04 & 78.85 & 89.42& 0.31 & 1.35 & 6.82  \\    
    & PixSfM& 79.98 & 87.84 & 94.52& \textbf{0.49} & \textbf{1.85} & \textbf{8.31} \\
    & DFSfM & 81.4 & 89.23 & 95.74 & 0.37 & 1.42 & 6.69\\
    & Ours & \textbf{83.37} & \textbf{90.74} & \textbf{96.58} & 0.38 & 1.49 & 7.01 \\
    \bottomrule
    \end{tabular}
    }
    \caption{\textbf{Comparison of sparse local features accompanied with our refinement and PixSfM, DetectorFree-SfM (DFSfM)}. Our method is compared with the baselines on the ETH3D dataset using accuracy and completeness metrics with different thresholds.}
    \label{tab: sparse_with_ours}
\end{table}

Our track refinement module (multi-view kernelized matching described in Sec.3.3) can also be applied to refine SfM models reconstructed using sparse detection and matching to further improve the robustness of SfM. We evaluate each SfM result on the ETH3D dataset, assessing accuracy and completeness following the ETH3D benchmark~\cite{eth3d_dataset}.

As shown in Tab.~\ref{tab: sparse_with_ours}, our refinement module consistently outperforms PixSfM and DFSfM in accuracy when accompanied by the same sparse detectors and matchers. Furthermore, our refinement module does not compromise the completeness much for higher accuracy.


\subsection{Semi-dense matching with our pipeline}

\begin{table}[ht!]
    \centering
    \resizebox{1.0\columnwidth}{!}{
    \setlength\tabcolsep{3pt} %
    \begin{tabular}{c|c|c|ccc|ccc}
    \toprule
    \multirow{2}{*}{Matcher} & 
    \multirow{2}{*}{Matching Preprocess} & \multirow{2}{*}{Refinement}  & \multicolumn{3}{c}{Accuracy~($\%$)} & \multicolumn{3}{c}{Completeness~($\%$)} \\
    \cmidrule(lr){4-6}
    \cmidrule(lr){7-9}
    &  & & 1cm       & 2cm       & 5cm       & 1cm       & 2cm       &  5cm \\ 
    \midrule
    \multirow{3}{*}{LoFTR} 
    & Quantization & DFSfM & 80.38 & 89.01  & 95.83 & 3.73 & 11.07 & 29.54  \\
    & Quanzization & Ours & 82.93 & 90.74 & 96.56 & 4.28 & 12.49 & 32.09 \\
    & Ours(GS-Extension) & Ours & \textbf{85.47} & \textbf{92.63} & \textbf{97.44} & \textbf{6.05} & \textbf{15.01} & \textbf{33.45} \\
    
    \bottomrule
    \end{tabular}
    }
    \caption{\textbf{Comparison of LoFTR(semi-dense matching) accompanied with matching process for track consistency and track refinement.} Our method is compared with the baselines on the ETH3D dataset using accuracy and completeness metrics with different thresholds.}
    
    \label{tab: semidense}
    \end{table}

\begin{table*}[h!]
    \centering
    \setlength\tabcolsep{3pt} %
    \resizebox{0.95\linewidth}{!}{
    \begin{tabular}{c| c c c c c c c c c c c c c |c} %
    
    \toprule 
        \multirow{2}*{Matching Design} & \multicolumn{13}{c|}{Scene Name} & \multirow{2}*{Avg.} \\
         \cline{2-14}
        & courtyard & delivery area & electro & facade & kicker & meadow & office & pipes & playground & relief & relief2 & terrace & terrains
        \\ 
        \hline
        \multicolumn{1}{c|}{} & \multicolumn{13}{c|}{\textit{Average Track Length}} &  \\ \hline

        Raw & 2.19 & 2.06 & 2.12 & 2.41 & 2.09 & 2.08 & 2.07 & 2.03  & 2.04 & 2.08 & 2.05  & 2.12 & 2.08 & 2.11\\  
       Quantized Matching (r=4) &3.43 & 3.11 & 3.37 & 4.82 & 2.90 & 2.86 & 2.91 & 2.89 & 3.05 & 3.21 & 3.13 & 3.19 & 3.14 & 3.23\\
       Track Extension via GS & 5.12 & 4.59 & 5.11 & 9.48 & 4.18 & 3.58 & 4.23 & 3.97 & 4.25 & 5.75 & 5.15 & 4.61 & 4.64 & \textbf{4.97}\\
               
        \bottomrule 
        \end{tabular}
    }
    \caption{Comparison of average track length accompanied with matching strategies (Quantization, Track Extension via GS) to obtain consistent tracks. 
    }
    \label{tab:track_length}
\end{table*}

We also apply our pipeline with semi-dense matcher, LoFTR, by replacing dense matchers (\eg DKM or RoMa) with LoFTR for building an initial SfM (without mutual verification). 
Specifically, to address pairwise matching's inconsistency, we compare our proposed Gaussian Splatting (GS) based track extension with quantization, and also evaluate our refinement module against DFSfM's refinement module.

As shown in Tab.~\ref{tab: semidense}, our GS based track extension not only provides superior accuracy but also improves completeness compared to quantization, highlighting its effectiveness. Furthermore, our refinement module consistently outperforms the one proposed in DFSfM, demonstrating the overall superiority of our pipeline when paired with a semi-dense matcher.

\subsection {Implementation Details}

As in PixSfM and DFSfM, we obtain match results from exhausitive pairs within a set of images. To obtain dense matching from image pairs, we resize images to $1162\times768$ for the ETH3D dataset and IMC 2021 dataset, while maintaining the original size of $840\times840$ for the Texture-Poor SfM dataset.

Since dense matching methods (\eg DKM and RoMa) are computationally heavier than detector-based or semi-dense approaches due to their pixel-wise matching process, we improve efficiency by obtaining coarse matches at a lower resolution ($560\times560$ for the ETH3D and IMC 2021, $420\times420$ for the Texture-Poor SfM dataset) in the global matching stage of RoMa, which dominates the runtime. We then refine the matches to the target resolution. 
Under this configuration, on the ETH3D dataset, RoMa achieves a runtime of 0.08 seconds per image pair on a single A6000 GPU. In comparison, DFSfM with LoFTR operates at a higher resolution ($1600\times1200$) and requires 0.2 seconds per pair. Our matching pipeline, which includes bidirectional matching with verification, runs in 0.17 seconds per pair, outperforming the LoFTR-based pipeline while maintaining a comparable matching speed.

\subsection{More Ablation Study}
\begin{table}[h]
    \centering
    \resizebox{1.0\columnwidth}{!}{
    \setlength\tabcolsep{8pt} %
    \begin{tabular}{ccccccc}
    \toprule
    \multirow{2}{*}{Refinement}    & \multicolumn{3}{c}{Camera Pose Estimation} & \multicolumn{3}{c}{Accu.(\%)~(\emph{Pipes})} \\ 
    \cmidrule(lr){2-4}
    \cmidrule(lr){5-7}
         & AUC@1$\degree$       & AUC@3$\degree$       & AUC@5$\degree$  & 1cm & 2cm & 5cm \\ 
    \midrule
    No Refine & 39.20 & 66.51 & 74.12 & 55.02 & 78.91 & 93.90 \\
    Iter 1 & 60.83 & 78.26 & 82.54 & 66.35 & 85.71 & 95.95 \\
    Track Extension via GS + Iter 2 & \textbf{60.92} & \textbf{78.41} & \textbf{82.63} & \textbf{74.75} & \textbf{92.06} & \textbf{97.98}\\
    \bottomrule
    \end{tabular}
    }
    \caption{\textbf{Ablation Study of Refinement Iterations.}
    On the ETH3D dataset, we quantitatively evaluate the impact of the number of refinement iterations. The AUC of pose error and accuracy of 3D points at different thresholds are reported.
    }
    \label{tab:sup_ablrefine}
\end{table}

In this section, we validate the effectiveness of our refinement pipeline by evaluating its performance on the reconstruction task, which involves recovering both camera poses and 3D point cloud. For evaluating point cloud accuracy, we use the \emph{Pipes} scene from the ETH3D dataset, as the scene provides ground-truth point clouds. We align reconstructed poses with the ground truth poses through COLMAP~\cite{colmap}, aligning 3D point as well, and evaluate the accuracy of point cloud following the ETH3D benchmark~\cite{eth3d_dataset}.

As shown in Tab.~\ref{tab:sup_ablrefine}, our pipeline improves the accuracy of both pose and point cloud through refinement. Note that the point cloud of \textit{No Refine} is triangulated from two-view non-quantized matching results (resulting in most tracks having a length of 2), based on coarse poses obtained from quantized matching as DFSfM. Thanks to track extension via Gaussian Splatting, we can feed more views into our multi-view kernelized refinement module, resulting in a slight improvement on camera pose accuracy, and a significant improvement on point cloud accuracy.

\subsection{Runtime Comparison}
We evaluate the runtime of our multi-view kernelized matching module against the multi-view matcher module in DFSfM\cite{detectorfreesfm} on the \textit{Pipes} scene, using the same SfM model for refinement. Our proposed refinement module costs a runtime increase about 22\%, from 35.3s to 43.2s on a single A6000 GPU, due to the added computational cost of the Gaussian Process and CNN Decoder.

\section{Details of Track extension via Gaussian Splatting}

We present here the detailed implementation of track extension via Gaussian Splatting.

\subsection{Training Gaussians}
\paragraph{Initialization.}
As described in Sec.3.2 in the main paper, 3D Gaussians are initialized based on the initial SfM point cloud. We set the initial Gaussians' position to the 3D point coordinates, opacity to 1, and rotation to the identity matrix. The scale parameter $S$ is determined using the method in Splatem~\cite{splatam}, where each Gaussian's radius is set such that it is projected as an one-pixel radius circle on the 2D image. Specifically, we compute depth values for each 3D point based on the camera poses of its keypoints. Then we select the maximum value across the depth values and divide it by the focal length as follows:

\begin{equation} S = \frac{D_{max}}{f} \end{equation} 

where $D_{max}$ represents the maximum depth value for the 3D point, and $f$ is the focal length.

\paragraph{Training details.}

Since the images of the dataset we used for evaluation consist of sparse views on a scene, we use training setup with Few-Shot Gaussian Splatting (FSGS)~\cite{fsgs} where it uses Proximity-guided Gaussian Unpooling~\cite{fsgs} for densifying Gaussians. For training loss, we only use photometric loss between rendered image and original image (excluding depth regularizer), because the estimated depth from a monocular depth estimation network can differ from the semi-dense depth provided by the initial SfM, potentially interfering with the visibility checks for the initialized Gaussians (SfM's 3D points). 

All images used in the SfM reconstruction are leveraged for training 3D Gaussians. Gaussians are densified every 500 iterations from the 1,500 iterations to 5000 iterations, with total optimization steps set to 6,000. The training time requires approximately 5 minutes on \textit{Pipes} scene in the ETH3D dataset on a single RTX 4090 GPU. Other training parameters follow those outlined in FSGS~\cite{fsgs}.

Unlike the standard training of GS~\cite{naive_gs} that requires over 10,000 iterations for high-quality rendering, our process focuses only on verifying the visibility of the initial SfM's 3D points. Thus, we do not need as many iterations as the general training for high quality image rendering~\cite{naive_gs, fsgs}.


\subsection{Track length analysis}

Tab.~\ref{tab:track_length} compared average track length.
The results show that our method significantly increases track length, which enables our refinement module to leverage more views, contributing to the higher accuracy reported in Tab.3 (1) of the main paper.


\section{Applying Gaussian Splatting}

\begin{table}[ht!]
\centering
\resizebox{\linewidth}{!}{
\begin{tabular}{c|cccc} \toprule
\multirow{2}{*}{SfM Initialization} & \multicolumn{4}{c}{Metrics}                                                                                                \\ \cline{2-5}
                               & L1$\downarrow$  & PSNR$\uparrow$ & SSIM$\uparrow$ & LPIPS$\downarrow$
                                  \\ \hline
SIFT                               & 0.0784 & 18.41 & 0.612 & 0.377 \\
LoFTR+DFSfM              & 0.0750 & 18.63 & 0.633 & 0.358 \\
RoMa+Ours                         & \textbf{0.0697} & \textbf{19.27} & \textbf{0.660} & \textbf{0.332} \\
\bottomrule
\end{tabular}
}
\caption{\textbf{Quantitative Comparison in the LLFF Dataset, with 3 Training Views.} 
Initialization with our method achieves the best performance in terms of rendering accuracy on all metrics.}
\label{tab:fsgs_result}
\end{table}

To demonstrate the utility of our framework, we use it to initialize 3D Gaussians for novel view synthesis in a few-shot setting. Starting from SfM points, we apply Few-Shot Gaussian Splatting (FSGS)~\cite{fsgs} on the LLFF dataset\cite{llff_dataset}. We compare the performance of 3D Gaussian models initialized using SIFT~\cite{sift} with COLMAP, DetectorFree-SfM (DFSfM) with LoFTR, and our framework with RoMa, using fixed camera poses and intrinsics provided by FSGS~\cite{fsgs} pipeline. For Gaussian initialization and training, we use only three views and render the original image size provided in the dataset.

For evaluation, test images are selected from the same scene in the LLFF dataset, excluding training images used for initialization and training. As shown in Tab.~\ref{tab:fsgs_result}, our method demonstrates significantly superior performance, thanks to dense and accurate initialized point cloud generated by our framework.

\section{Limitations and Future works}
 A key limitation of our framework is that track extension via Gaussian Splatting struggles in scenes with high photometric variation or transient occlusions. As future work, our pipeline can be extended with more advanced Gaussian Splatting models~\cite{wildgaussians, wild2, wild3, wild4}, to better handle occlusions and appearance changes, allowing our framework to perform robustly in a wider range of scenarios.


\end{document}